%% file: iclr2023_conference.tex
\documentclass{article} % For LaTeX2e
\usepackage{iclr2023_conference,times}

\input{math_commands.tex}

\usepackage{hyperref}
\usepackage{url}

\usepackage[utf8]{inputenc} % allow utf-8 input
\usepackage[T1]{fontenc}    % use 8-bit T1 fonts
\usepackage{hyperref}       % hyperlinks
\usepackage{url}            % simple URL typesetting
\usepackage{booktabs}       % professional-quality tables
\usepackage{amsfonts}       % blackboard math symbols
\usepackage{nicefrac}       % compact symbols for 1/2, etc.
\usepackage{microtype}      % microtypography
\usepackage{xcolor}         % colors

\usepackage{multirow}
\usepackage{subfigure}
\usepackage{amsmath,amssymb,graphicx,amsthm,xparse, color, mathrsfs} 
\usepackage{algpseudocode}
\usepackage{bbm}

\usepackage{amsthm}
\theoremstyle{definition}

\newcommand{\etal}[0]{\textit{et al.}}

\title{ A Multimodal Transformer: Fusing Clinical Notes with Structured EHR Data for Interpretable In-Hospital Mortality Prediction }

\iclrfinalcopy
\author{Weimin Lyu\textsuperscript{\textnormal{1}}, Xinyu Dong\textsuperscript{\textnormal{1}}, Rachel Wong\textsuperscript{\textnormal{1}}, Songzhu Zheng\textsuperscript{\textnormal{1}}, Kayley Abell-Hart\textsuperscript{\textnormal{1}}, \And
Fusheng Wang\textsuperscript{\textnormal{1}}, Chao Chen\textsuperscript{\textnormal{1}} \\ \\
\textsuperscript{1} Stony Brook University, NY, USA}

\begin{document}

\maketitle

\begin{abstract}
Deep-learning-based clinical decision support using structured electronic health records (EHR) has been an active research area for predicting risks of mortality and diseases. Meanwhile, large amounts of narrative clinical notes provide complementary information, but are often not integrated into predictive models. In this paper, we provide a novel multimodal transformer to fuse clinical notes and structured EHR data for better prediction of in-hospital mortality. To improve interpretability, we propose an integrated gradients (IG) method to select important words in clinical notes and discover the critical structured EHR features with Shapley values. These important  words and clinical features are visualized to assist with interpretation of the prediction outcomes. We also investigate the significance of domain adaptive pretraining and task adaptive fine-tuning on the Clinical BERT, which is used to learn the representations of clinical notes. Experiments demonstrated that our model outperforms other methods (AUCPR: 0.538, AUCROC: 0.877, F1:0.490). 
\end{abstract}

\section{Introduction}

Electronic health record (EHR) systems are widely used in the United States\citep{henry2016adoption} and the large amount of EHR data generated provides an opportunity for machine learning based predictive modeling to improve clinical decision support. In particular, deep learning based techniques\citep{dalal2021evaluation, schwartz2021clinician}, have been used for prediction of in-hospital mortality\citep{kong2020using, li2021prediction}, diagnoses\citep{dong2019machine, yang2021leverage}, length of stay\citep{cai2016real}, readmission\citep{teo2021current}.

EHRs include structured data and clinical notes, which are often unstructured\citep{zheng2017effective}. Structured clinical variables, such as the vital signals (e.g., heart rate, respiration rate, temperature, and blood pressure), can be easily converted to time series data and are well explored by researchers\citep{sheikhalishahi2020benchmarking, rocheteau2021predicting, si2021deep}. For example, Harutyunyan \etal \citep{harutyunyan2019multitask} establishes a benchmark on how to pre-process the MIMIC III dataset, and proposed various baselines for different tasks, e.g., Logistic Regression, Random Forest, Recurrent Neural Network (RNN), Long Short-Term Memory (LSTM), and Convolutional Neural Network (CNN). Dong \etal \citep{dong2019machine} develops a machine learning based opioid overdose prediction method with different clinical variables. Unstructured clinical notes are often in free narrative form, but  contain complementary and rich contextual information, such as a patient’s symptoms, disease course and treatment\citep{lee2020biobert}. Though the normally pre-trained natural language model Bidirectional Encoder Representations from Transformers (BERT)\citep{devlin2019bert} cannot handle specific clinical notes, there are different variants of BERTs\citep{harutyunyan2019multitask, gururangan2020don, alsentzer2019publicly}, that are pretrained on biomedical and clinical data, which can better handle  clinical notes. For example, Clinical BERT\citep{huang2019clinicalbert} pretrains the BERT using MIMIC III clinical notes with masked language modeling (MLM) and next sentence prediction (NSP), to predict hospital readmission. BEHRT\citep{li2020behrt} incorporates age and position information when modeling the clinical notes. BioBERT\citep{lee2020biobert} is pretrained on biomedical notes like PubMed abstracts and PubMed Central full-text articles to significantly improve biomedical text mining task performance. BioRoberta\citep{gururangan2020don} points out that in-domain pretraining leads to performance gains. Clinical BERT\citep{huang2019clinicalbert} uses a domain-specific model to improve performance on three common clinical NLP tasks. BLURB\citep{gu2021domain} benchmark is a recent work that released state-of-the-art pretrained and task-specific models for the community. Despite these advances, how to  leverage and interpret the information included in unstructured clinical notes remains a challenging problem.

Multimodal fusion is a promising direction to address the aforementioned challenge. However, naively concatenating features from different modalities might result in worse performances\citep{ramachandram2017deep}. It is challenging to embed data from the structured clinical variables and unstructured clinical notes  together because they are two totally different domains. Si \etal \citep{si2021deep} provides a comprehensive survey on deep representation learning from single and multiple domains of EHR data. Some works merely extract features from structured and unstructured data separately, and then concatenate the two features\citep{zhang2020combining}. For example, Khadanga \etal \citep{khadanga2019using} extracts clinical notes with convolutional neural networks and incorporates time series data to improve the performance. Deznabi \etal \citep{deznabi2021predicting} models two modalities with Long short-term memory (LSTM) and BERT. Yang \etal \citep{yang2021leverage} implements Multimodal Adaptation Gate (MAG)\citep{rahman2020integrating} techniques to best utilize information from two modalities. Teixeira \etal \citep{teixeira2017evaluating} tested different combinations of several different modalities. Huang \etal \citep{huang2020fusion} discuss fusion strategies of structured data and imaging data. However, these methods naively concatenate without considering complexity of modality and time information. While transformers are gaining popularity for use in different domains,  there is limited work using  transformers on EHR based predictive modeling.

In this paper, we propose a multimodal transformer to fuse time series data from clinical variables with textual information from clinical notes to boost performance of  in-hospital mortality prediction. We leverage clinical notes to provide auxiliary information by adjusting the representation from two modalities to a sharable space across different times, then jointly learn the representation from two modalities. Further, we implement the transformer on the time series EHR data to fully consider the information across time, combined with the fine-tuned Clinical BERT model, which is a novel application of EHR feature representation learning. We also show that pre-training of various BERT models results in different prediction ability with regard to clinical tasks, and the BERT models fine-tuned on the specific in-hospital prediction task brings further performance improvements. Furthermore, we extend our fine-tuned Clinical BERT model with the integrated gradients (IG) method to interpret and visualize the important words in clinical notes. Our results demonstrate that by leveraging the clinical notes, our proposed Multimodal Transformer provides highly promising prediction results (with AUCPR: 0.538, AUCROC: 0.877, F1:0.490). \textit{To our best knowledge, our Multimodal Transformer is the first work utilizing the transformer block to fuse clinical notes information and clinical variable information, while including time series EHR data.}

\section{Methods}

\subsection{Study Dataset}

We extract inpatient EHR data from the Medical Information Mart for Intensive Care (MIMIC-III) dataset\citep{johnson2016mimic}. The clinical variables are pre-processed as time series signals from ICU instruments following Harutyunyan \etal \citep{harutyunyan2019multitask} benchmark setup. Seventeen clinical variables remained after preprocessing: capillary refill rate; diastolic blood pressure; fraction inspired oxygen; the eye opening, motor response, verbal response, and total value of the Glasgow Coma Scale; glucose; heart rate; height; mean blood pressure; oxygen saturation; respiratory rate; systolic blood pressure; temperature; weight; and pH. 

For the clinical notes, similar to the setup from Khadanga \etal\citep{khadanga2019using}, we extract notes from the NOTEEVENTS.csv file, and remove all clinical notes that do not have any chart time associated and remove patients that do not have any clinical notes. While Khadanga \etal kept only the first visit for each patient, we treat every visit as a single sample. Therefore, in the following paper, we use ‘patient’ to indicate ‘visit’. After the above data processing, our dataset for in-hospital mortality prediction contains 14068 training samples, 3086 validation samples, and 3107 test samples. Due to this step, our results are not directly comparable to the benchmark from Harutyuanyan \etal \citep{harutyunyan2019multitask}.

\begin{figure}[ht]
\centering
\vspace{-.2in}
\includegraphics[width=13.5cm]{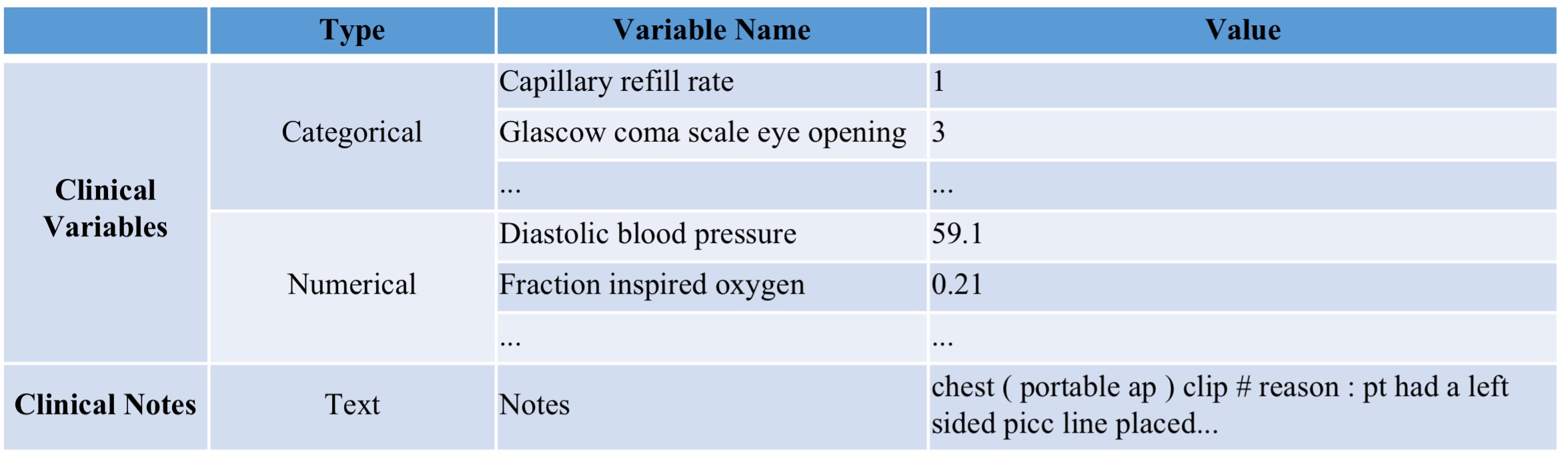}
\caption{An example of MIMIC III EHR data for ICU patients, containing two modalities: clinical variables and clinical notes. The clinical variables can be further split into categorical and numerical variables, while the clinical notes are domain specific text.}
\label{fig:ehr_illustration}
\end{figure}

\subsection{Single Model Embedding}

We aim to predict in-hospital mortality with multi-modal EHR data. First, we  process two single modalities (clinical notes and clinical variables) separately to obtain the initialized embeddings from the raw data. We introduce Notes Embedding and Variables Embedding to achieve the initialized single modality embedding.

\textbf{Clinical Notes Embedding.} Though BERT models dominate increasing numbers of domains in NLP, BERT-based models do not necessarily offer strong clinical text mining ability with regard to a specific clinical task. Rather, the power of BERT-family models relies on domain adaptive pretraining and task adaptive fine-tuning. Pre-training on proper clinical biomedical corpora enables the BERT-based model to better learn clinical contextual meaning representations, and fine-tuning on downstream tasks can further boost this ability and establish a specialized Clinical BERT model. To illustrate this point, we compare the representation ability of four different BERT models (Table \ref{tab:pretrain}) using only single clinical notes modality, namely BERT\citep{devlin2019bert}, BioBERT\citep{lee2020biobert}, BioRoBERTa\citep{gururangan2020don}, Clinical BERT\citep{huang2019clinicalbert}, pertained on four types of corpora, respectively English Wikipedia / BooksCorpus, PubMed Abstracts / PMC Full-text articles (initialized from BERT), S2ORC,33 and entire MIMIC III notes (initialized from BioBERT). Detailed results are shown in Table \ref{tab:ehr_ablation}.

% Table 1
\begin{table}[]
\caption{Four BERT models and their respective corpora used for pretraining. Initialized model indicates the starting point before pretraining.}
\label{tab:pretrain}
\centering
\begin{tabular}{c|c|c|c}
\hline
\textbf{Pretrained Model} & \textbf{Pretraining Corpora}             & \textbf{Initialized Model} & \textbf{Domain} \\ \hline
\textbf{BERT}             & English Wikipedia, BooksCorpus           &                            & General         \\
\textbf{BioRoBERTa}       & S2ORC                                    & RoBERTa                    & Biomedical      \\
\textbf{BioBERT}          & PubMed Abstracts, PMC Full-text articles & BERT                       & Biomedical      \\
\textbf{Clinical BERT}    & MIMIC notes                              & BioBERT                    & Biomedical      \\ \hline
\end{tabular}
\vspace{-.1in}
\end{table}

We select Clinical BERT\citep{huang2019clinicalbert} as our pre-trained language model since it is a more proper domain-specific model trained on all MIMIC-III clinical notes. We further fine-tune the Clinical BERT with the in-hospital mortality prediction task on MIMIC-III, called MIMIC BERT (MBERT), which enables the Clinical BERT to learn better clinical specific contextual embeddings on specific MIMIC data. For each patient, we extract an embedding of the clinical notes for every associated hour to represent the clinical notes data with time information. In the following experiments, we freeze the weights of MBERT when extracting unstructured clinical notes embeddings in multimodal transformer, since the MBERT already preserves a good clinical meaning representation.

% We aim to predict in-hospital mortality with multi-modal EHR data. First, we  process two single modalities (clinical notes and clinical variables) separately to obtain the initialized embeddings from the raw data. We introduce Notes Embedding and Variables Embedding to achieve the initialized single modality embedding.
% Clinical Notes Embedding. We compare the representation ability of 4 BERTs (details in Results Section) using only single clinical notes modality, and select Clinical BERT\citep{huang2019clinicalbert} as our pre-trained language model since it is a more proper domain-specific model trained on all MIMIC-III clinical notes. We further fine-tune the Clinical BERT with the in-hospital mortality prediction task on MIMIC-III, called MBERT, which enables the Clinical BERT to learn better clinical specific contextual embeddings on specific MIMIC data. For each patient, we extract an embedding of the clinical notes for every associated hour to represent the clinical notes data with time information. 

\begin{figure}[ht]
\centering
\includegraphics[width=13.5cm]{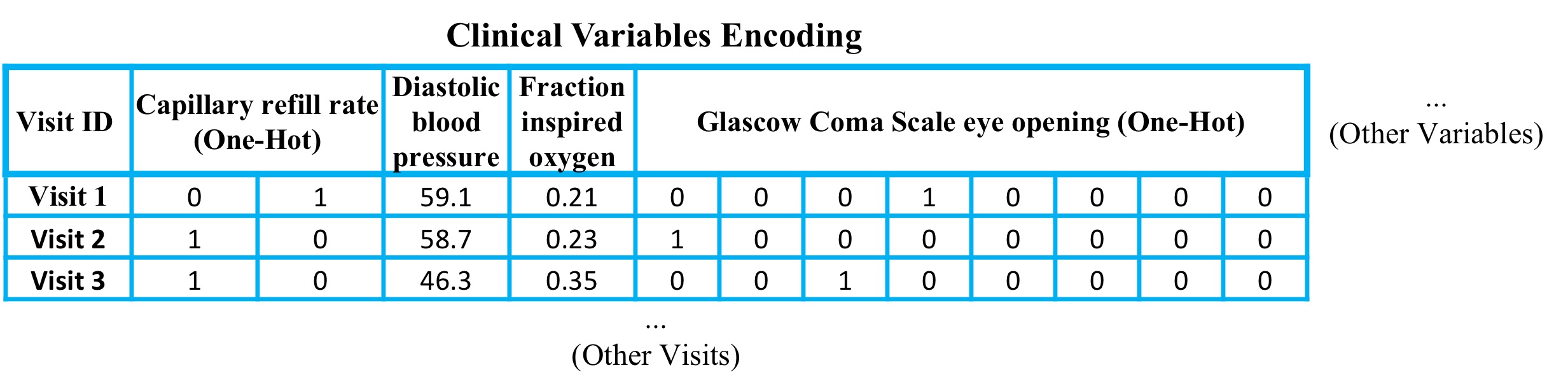}
\caption{An illustration of Clinical Variables Encoding.}
\label{fig:ehr_onehot}
\end{figure}

\textbf{Clinical Variables Embedding.} Given that clinical variables contain numerical and categorical data, we apply one-hot encoding to the clinical variables, illustrated in Figure \ref{fig:ehr_onehot}. Following Harutyunyan's setup, the 17 clinical variables are embedded to a 76-dimension time series embedding after the one-hot encoding. The categorical variables are converted to one-hot vectors while the numerical variables are converted to a single continuous value.

For a formal mathematical representation, we denote the clinical notes data as $X_{notes} \in \R^{L\times D_1}$, where $L$ represents the length of ICU stay counted by hours, and $D_1$ represents the maximum length of clinical notes. And we denote the clinical variables data as $X_{ts} \in \R^{L\times D_2}$, where $D_2$ represents the number of variables. The clinical notes are embedded with Fine-tuned Clinical BERT (MBERT), as $E_{notes}=MBERT(X_{notes})$, and the clinical variables are embedded as $E_{ts}=Variable\_Encoding(X_{ts})$.

\subsection{Multimodal Embedding }

We introduce a transformer to integrate two different modalities. Specifically, we introduce three encoders inside the transformer block: Notes Encoder and Time Series (TS) Encoder for clinical notes and clinical variables modalities separately, and Multimodal (MM) Encoder to fuse two modalities while projecting them into a shared space:

\textbf{Encoders.} (1) Notes Encoder. Given that the clinical notes embedding $E_{notes}$ is already well presented, we only use a single linear layer to project $E_{notes}$ to a universal space. (2) Time Series (TS) Encoder. Since the clinical variables embedding contains  simple numerical and categorical information, we also use linear layers to project $E_{ts}$ to a universal space. (3) Multimodal (MM) Encoder. We do not simply concatenate the clinical notes and clinical variables because the two modalities are  conceptually different. We use a Multimodal Encoder to compact the two different modalities into a universal space before we feed them into the transformer model, so that the information from clinical notes and clinical variables can be jointly learned. The formal mathematical representation is as follows:

\begin{equation}\label{eq:ehr_representation}
\begin{array}{lll}
I_{notes}=Encoder_{notes}(E_{notes}) \\
I_{ts}=Encoder_{ts}(E_{ts})\\
I_{MM}=Encoder_{MM}(E_{notes} \oplus E_{ts})
\end{array}
\end{equation}

where $\oplus$ denotes concatenate operation, $I_{notes}\in \R ^{L\times D_3}$, $I_{ts}\in \R ^{L\times D_4}$, $I_{MM}\in \R ^{L\times D_5}$ denotes outputs from associate encoders, and $D_3,D_4,D_5$ represent the corresponding embedding dimension.

\textbf{Transformer.} Our transformer block is the key to handle time series embeddings and integrate knowledge. The transformer is a popular model developed for natural language processing (NLP), and has emerged as a promising tool in other domains. In LSTM, if the time sequence is too long, then when the information is passed to the final timestamp, the model forgets the information in the earlier timestamps. The powerful attention mechanism in the transformer enables the model to better leverage the information from all the timestamps. However, more research is needed to determine how best to apply the transformer in clinical tasks, especially when using multimodal data. We successfully implement the transformer in our study to show the capability of the transformer model.

\begin{figure}[ht]
\centering
\includegraphics[width=13.5cm]{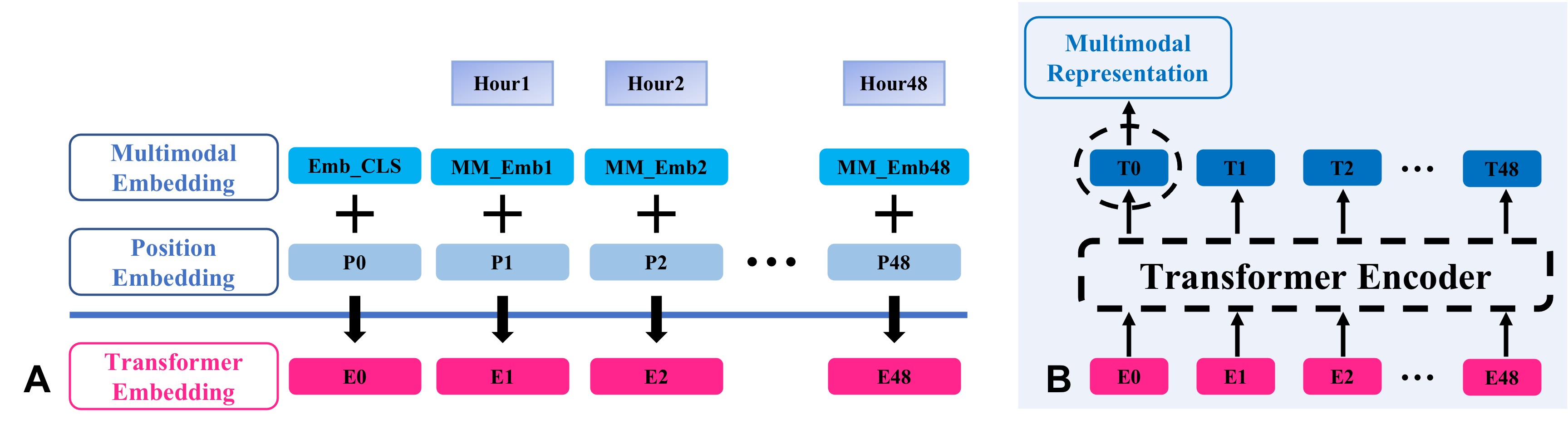}
\caption{An illustration of the Transformer architecture. A. Adding the position embedding to consider time information. B. Details of the transformer block. The fused transformer embedding is fed into the transformer encoder, and we only select the ‘T0’ token as the final multimodal representation.}
\label{fig:ehr_transformer2}
\end{figure}

\begin{figure}[ht]
\centering
\includegraphics[width=0.4\textwidth]{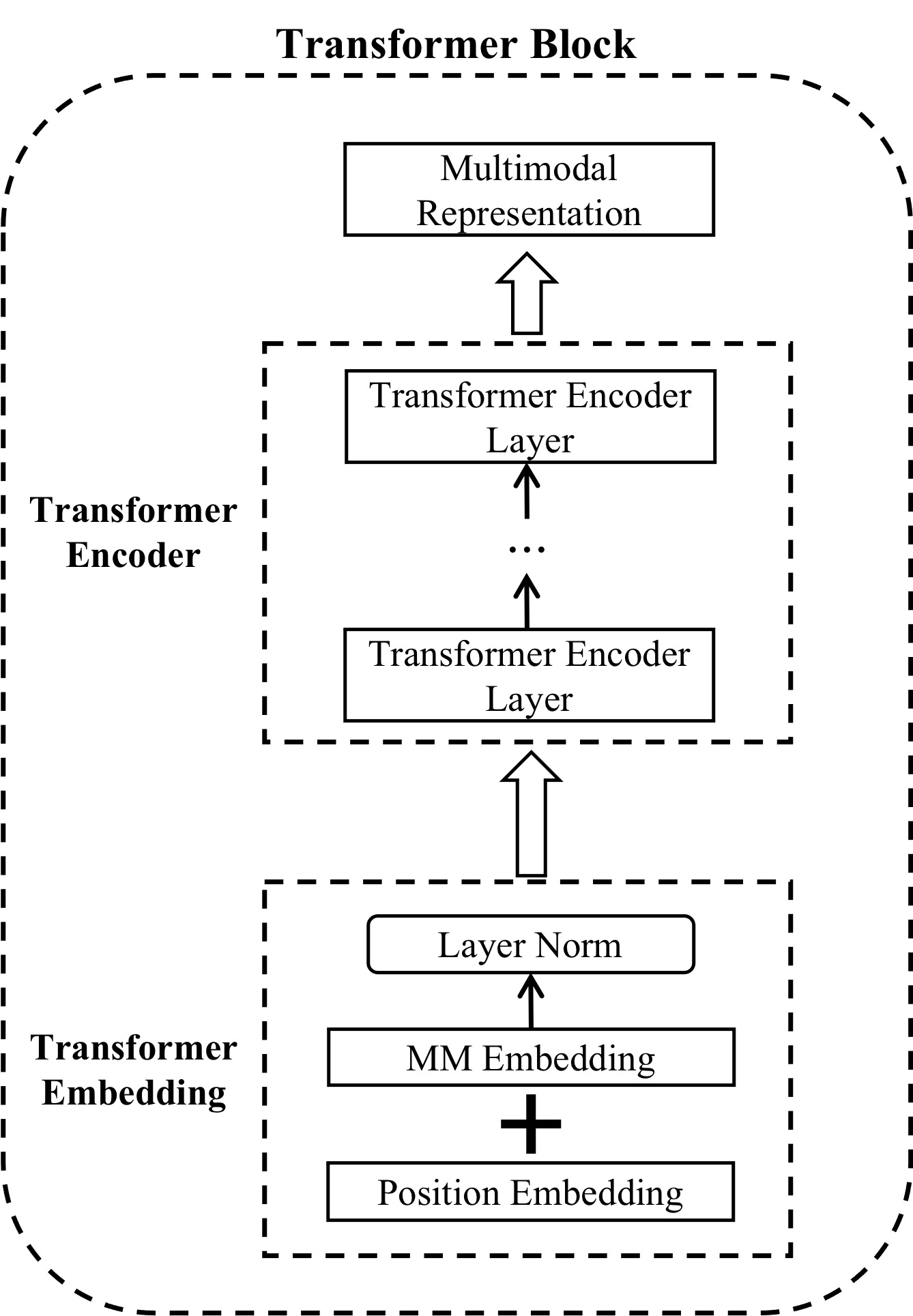}
\caption{An illustration of the Transformer block.}
\label{fig:ehr_transformer1}
\end{figure}

In the NLP context, if one sentence has 48 tokens after tokenization, then the input of the NLP model would be a 48-length sequence. Similarly, in the clinical time series context, we treat each hour as one token. Since we consider the first 48 hours in the ICU, there are 48 `tokens' for one patient. In Figure \ref{fig:ehr_transformer2}, the multimodal embedding of one patient is shown. The position embedding encodes the time information. In this way, the transformer block is able to consider information from all the time sequences when learning the representations. We use sinusoidal positional embedding in our model. Similar to the NLP techniques, we insert the `CLS' token at the beginning of the time sequences and use the T0 as the final multimodal representation. Figure \ref{fig:ehr_transformer1} illustrates the detailed architecture of the whole transformer block, with a formal mathematical presentation:

$$I_{Multimodal} = Transformer(I_{MM})$$

Then we concatenate the multimodal representations $I_{Multimodal}$ and notes embedding $E_{notes}$ to get the final prediction: 

$$Pred = MLP(I_{Multimodal} \oplus E_{notes})$$

\subsection{Overview Architecture}

\begin{figure}[ht]
\centering
\includegraphics[width=\textwidth]{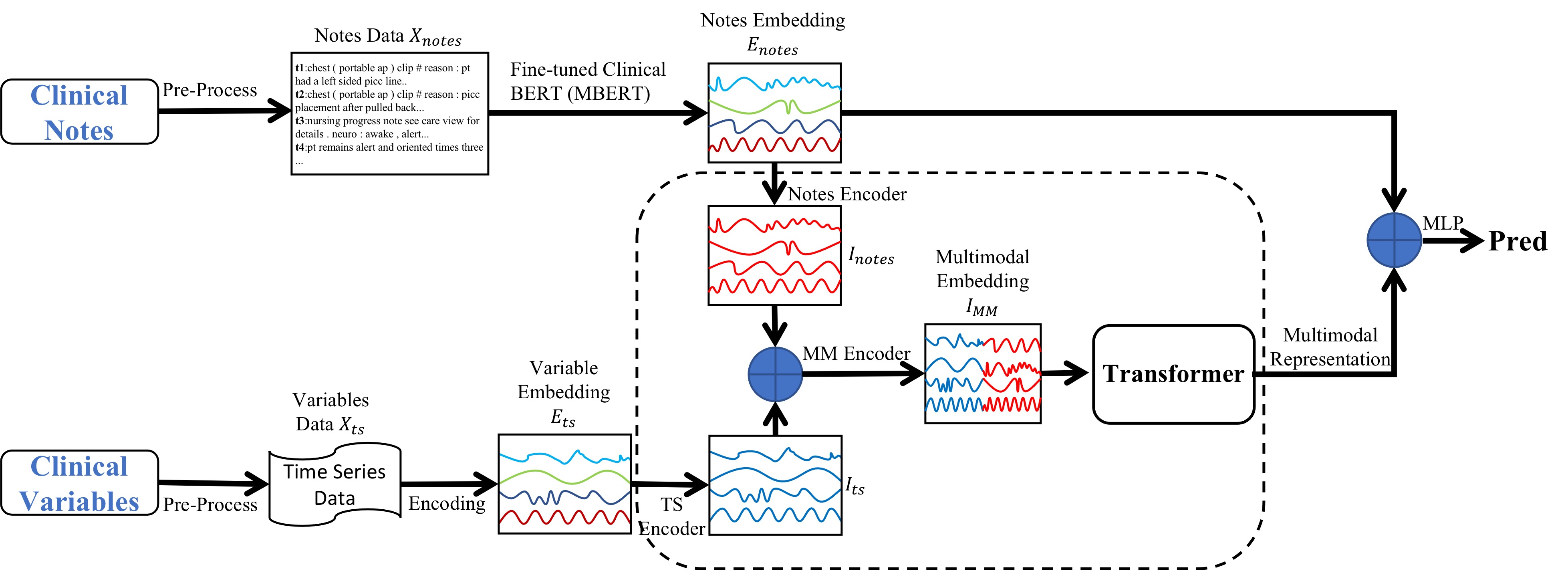}
\caption{The overview architecture of our proposed Multimodal Transformer. }
\label{fig:ehr_architecture}
\end{figure}

The overall architecture of our Multimodal Transformer is shown in Figure \ref{fig:ehr_architecture}.

\textbf{Implementation.} In our experiment, a rectified linear unit (ReLU) function is used as the non-linear projection function across different layers to prevent vanishing gradient and sparse activation problems. The sigmoid function is applied in the last layer. We use cross entropy loss and L2 regularization as the loss function and the Adam optimization to minimize the loss. We use Python Programming Language (Version3.8). Models are implemented with Python Pytorch\citep{paszke2019pytorch} and HuggingFace Transformers\citep{wolf2019huggingface}. The training was performed on an NVIDIA RTX A5000 (24GB RAM).

\section{Results}
\subsection{Prediction Results Analysis}
We predict in-hospital mortality based on the first 48 hours of an ICU stay, which is a binary classification task. We use the same train-test setting defined in the benchmark\citep{harutyunyan2019multitask} with  15\% of the training data as a validation set, and similar to Khadanga \etal\citep{khadanga2019using}, we remove all clinical notes that do not have any chart time associated and patients that do not have any clinical notes. The statistics on the post-processed data are shown in Table \ref{tab:ehr_stats}.

\begin{table}[]
\centering
\caption{Statistics  of the post-processed MIMIC III data for the in-hospital mortality prediction task.}
\label{tab:ehr_stats}
\begin{tabular}{c|c|c|c}
\hline
                  & \textbf{Train} & \textbf{Validation} & \textbf{Test} \\ \hline
\textbf{Negative} & 12216          & 2682                & 2748          \\ \cline{1-1}
\textbf{Positive} & 1852           & 404                 & 359           \\ \hline
\textbf{Total}    & 14068          & 3086                & 3107          \\ \hline
\end{tabular}
\end{table}

To comprehensively evaluate the performance of our model, we compute the metrics AUCROC, AUCPR, and F1. As the dataset is imbalanced, other metrics such as accuracy, recall, and precision may be misleading. We run all experiments five times with different initialization and report the mean and standard deviation of the results.

Results in Table \ref{tab:ehr_results} demonstrate that our models outperform other methods in classifying in-hospital mortality. We achieve an AUCPR score of 0.538, an AUCROC score of 0.877, and an F1 score of 0.490.

%table 3
\begin{table}[ht]
\centering
\caption{Experiment Results of different methods on MIMIC III In-Hospital Mortality Prediction Task.}
\label{tab:ehr_results}
\resizebox{\columnwidth}{!}{ %% resize table to fix the file
\begin{tabular}{c|c|c|c|c}
\hline
                                         & \textbf{Prediction Model}     & \textbf{AUCPR}                              & \textbf{AUCROC}                             & \textbf{F1}                                 \\ \hline
\multirow{2}{*}{\textbf{Only Variables}} & LSTM                          & 0.460($\pm$0.013)              & 0.821($\pm$0.006)              & 0.392($\pm$0.038)              \\
                                         & Transformer                   & 0.473($\pm$0.011)              & 0.827($\pm$0.005)              & 0.406($\pm$0.025)              \\ \hline
\textbf{Only Notes}                      & MBERT                         & 0.482($\pm$0.012)              & 0.851($\pm$0.005)              & 0.382($\pm$0.079)              \\ \hline
\multirow{2}{*}{\textbf{Fusion}}         & MBERT+LSTM                    & 0.508($\pm$0.002)              & 0.859($\pm$0.001)              & 0.478($\pm$0.023)              \\
                                         & Multimodal Transformer (Ours) & \textbf{0.538($\pm$0.004)} & \textbf{0.877($\pm$0.001)} & \textbf{0.490($\pm$0.036)} \\ \hline
\end{tabular}
}
\end{table}

% table 4
\begin{table}[]
\centering
\caption{Experiments on various Pre-trained and Fine-tuned BERTs. Use only MIMIC III clinical notes for in-hospital mortality prediction without considering clinical variables information. Freeze indicates only training the final classifier while keeping the BERT models unchanged. Fine-tuned indicates fine-tuning the BERTs for the in-hospital mortality downstream task.}
\label{tab:ehr_ablation}
\resizebox{\columnwidth}{!}{ %% resize table to fix the file
\begin{tabular}{|c|ccc|ccc|}
\hline
                       & \multicolumn{1}{c|}{\textbf{AUCPR}}          & \multicolumn{1}{c|}{\textbf{AUCROC}}         & \textbf{F1}     & \multicolumn{1}{c|}{\textbf{AUCPR}}          & \multicolumn{1}{c|}{AUCROC}                  & F1                      \\ \hline
                       & \multicolumn{3}{c|}{\textbf{Freeze}}                                                                          & \multicolumn{3}{c|}{\textbf{Fine-tuned}}                                                                              \\ \hline
\textbf{BERT}          & \multicolumn{1}{c|}{0.182($\pm$0.016)}          & \multicolumn{1}{c|}{0.649($\pm$0.020)}          & 0($\pm$0)          & \multicolumn{1}{c|}{0.417($\pm$0.023)}          & \multicolumn{1}{c|}{0.829($\pm$0.005)}          & 0.342($\pm$0.054)          \\ \hline
\textbf{BioRoBERTa}    & \multicolumn{1}{c|}{0.182($\pm$0.013)}          & \multicolumn{1}{c|}{0.661($\pm$0.016)}          & 0($\pm$0)          & \multicolumn{1}{c|}{0.455($\pm$0.010)}          & \multicolumn{1}{c|}{0.841($\pm$0.005)}          & \textbf{0.419($\pm$0.044)} \\ \hline
\textbf{BioBERT}       & \multicolumn{1}{c|}{0.191($\pm$0.005)}          & \multicolumn{1}{c|}{0.664($\pm$0.011)}          & 0($\pm$0)          & \multicolumn{1}{c|}{0.444($\pm$0.027)}          & \multicolumn{1}{c|}{0.843($\pm$0.006)}          & 0.377($\pm$0.045)          \\ \hline
\textbf{Clinical BERT} & \multicolumn{1}{c|}{\textbf{0.265($\pm$0.006)}} & \multicolumn{1}{c|}{\textbf{0.731($\pm$0.004)}} & 0($\pm$0) & \multicolumn{1}{c|}{\textbf{0.482($\pm$0.012)}} & \multicolumn{1}{c|}{\textbf{0.851($\pm$0.005)}} & 0.382($\pm$0.079)          \\ \hline
\end{tabular}
}
\end{table}

In the following section, we first investigate variants of BERT models with regard to pretraining and fine-tuning. Then, we visualize the important words in clinical notes by Integrated Gradient. Finally, we analyze the important clinical variables with the Shapley value\citep{lundberg2017unified}.

\subsection{Domain adaptive pretraining and task adaptive fine-tuning on BERT models}

In order to show the importance of domain adaptive pretraining and task adaptive fine-tuning in BERTs, we conduct an ablation study with only the pretrained models versus with fine-tuning using a single modality - clinical notes. The results are shown in Table \ref{tab:ehr_ablation}. As expected, the general-purpose ‘BERT’ achieves the poorest result, whereas the MBERT achieves the best performance. These experiments suggest that clinical notes with proper trained language model are able to provide helpful information in clinical tasks, which enables deep learning techniques to leverage rich textual information to better understand the patient situation.

\subsection{Clinical Notes Visualization and Interpretation}
\begin{figure}[ht]
\centering
\includegraphics[width=0.8\textwidth]{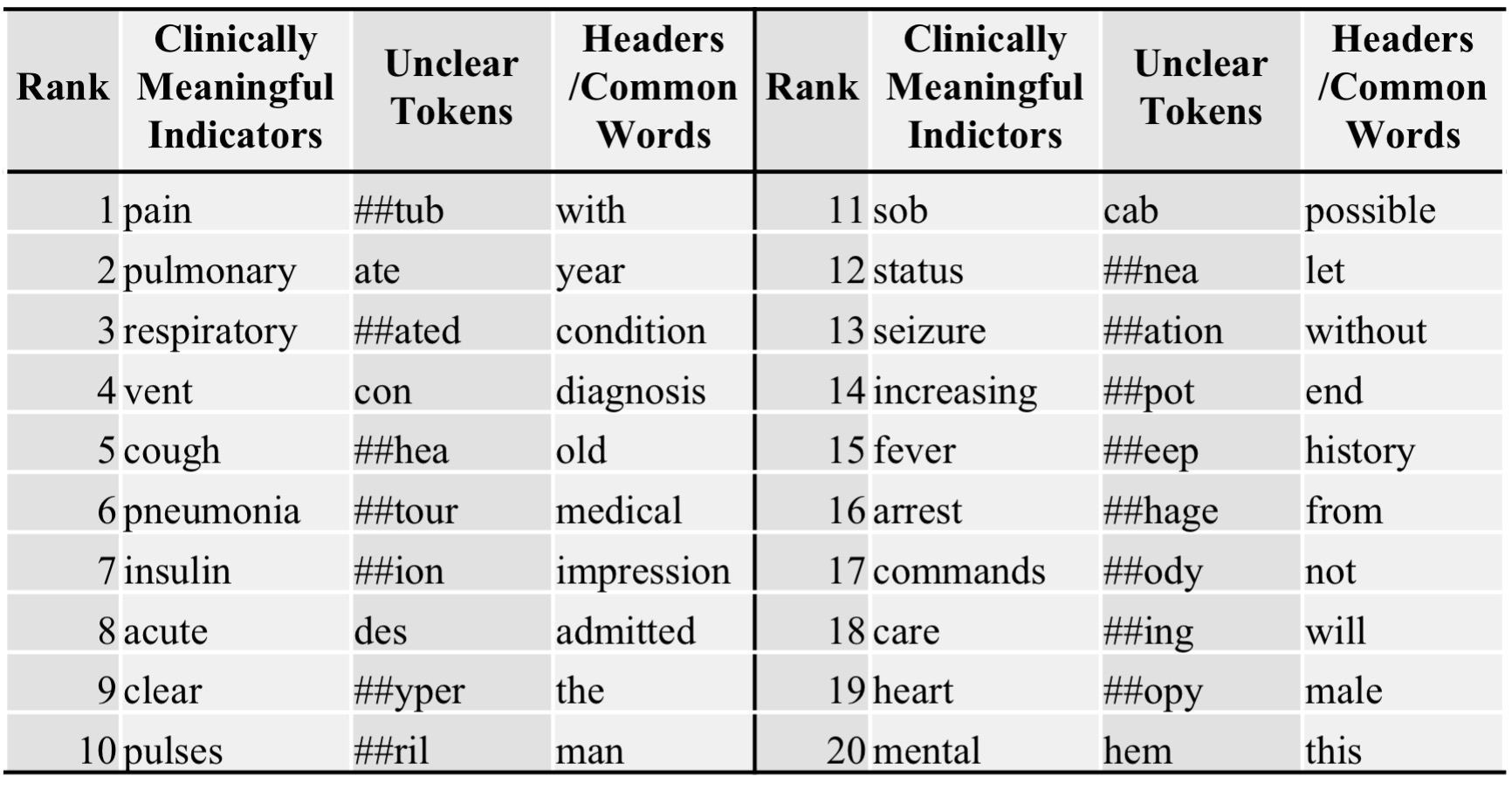}
\caption{Top 20 Features Integrated Gradient Values Sorted for interpretability. The results come from patients in test data.}
\label{fig:ehr_wordig}
\end{figure}

To provide an interpretation for the clinical notes and to better visualize the information, we evaluated the words that were important for prediction in our MBERT model using Integrated Gradients (IG)\citep{sundararajan2017axiomatic}. We apply the IG method to study the problem of attributing the prediction of a deep network to its input features, as an attempt towards explaining individual predictions. IG is computed based on the gradient of the prediction outputs considering the input words. Higher IG values indicate that a word is more important to the model’s prediction, while smaller IG values indicate that a word is less important. We compute the IG value of all tokens in the clinical notes for all patients in the test data, and list the tokens with the highest IG values. Note that due to the BERT tokenization mechanism, the inputs are tokens instead of words. For example, the phrase ``the patient has been extubated'' would be tokenized to ``the patient has been ex $\#\#$tub $\#\#$ated'' as the input. To make the results more readable, we remove all the numbers,  tokens that only have one or two characters, and separators in post processing. The tokens and their IG values are evaluated by a clinician for their clinical meaningfulness for mortality prediction. The tokens are sorted by those that are ``Clinically Meaningful Indicators'' of symptoms, prognosis, or care; ``Unclear Tokens'' which are difficult to attribute a single meaning to; and Headers/Common Words that are parts of structured notes or ubiquitously words used in medical notes, illustrated in Figure \ref{fig:ehr_wordig}.

Several words with high IG values appear to be parts of structured headers, such as “medical condition,” “diagnosis” or “impression,” so are categorized separately from text that was unstructured. Additional words that are used ubiquitously in clinical notes, such as “year,” “old,” and “with” are also categorized separately as they were less likely to distinguish prognostic differences. Evaluating the top 20 clinically meaningful indicators that are important for mortality prediction, there are some interesting observations for clinical interpretation. “Pain”, which is the indicator most important for prediction, is a common symptom in ICU care and can correlate with disease severity or disability. Indicators 2-6 correspond to pulmonary pathology, and the attribution of high importance to these indicators is in line with severity of respiratory illness and the need for ICU level care such as mechanical ventilation. Other indicators, such as “fever” or “seizure”, are manifestations of acute illness, which could also have prognostic significance in predicting mortality. Clinical indicators such as “status,” “commands,” “mental,” and “agitation” corresponded to mental status, and as delirium is associated with worse prognosis, it is not surprising that these indicators have prognostic importance in prediction.35,36 Additional words such as “care” had multiple contexts when reviewing the notes; phrases such as “plan of care” or “resp care” are often used as headers, but used in other contexts it could be interpreted as a poor prognostic signal (e.g. “withdrawal of care”)  or a favorable prognostic signal (e.g. “ response to care”).

\begin{figure}[ht]
\centering
\includegraphics[width=0.8\textwidth]{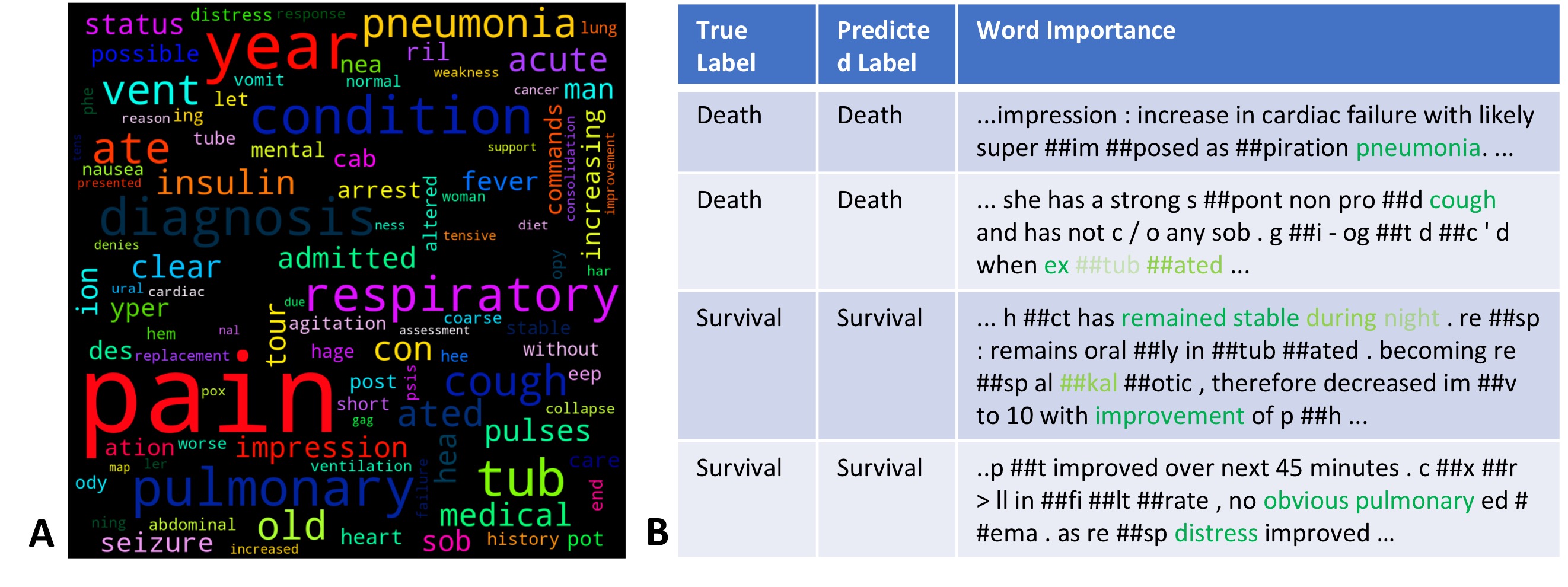}
\caption{A: Word Cloud for clinical tokens with high IG values. Larger font indicates the word is more likely to appear as a top-ten IG value in clinical notes. B: Illustration of word importance in clinical notes based on IG value. The darker green color indicates the words that are more important (higher IG value) to the prediction, while the black color is background color.}
\label{fig:ehr_word_cloud}
\end{figure}

Figure \ref{fig:ehr_word_cloud}.A is the word cloud visualization of the top 200 important words. We select the top 10 words with highest IG in every note, and compute all the notes. Says there are 10000 notes, then there would be 10*10000 top words (repeatable), and we compute the frequency of each unique word. The font size reflects the frequency. Figure \ref{fig:ehr_word_cloud}.B is a demo illustration of word importance among the clinical notes.

\subsection{Clinical Variables Feature Analysis}
\begin{figure}[ht]
\centering
\includegraphics[width=0.8\textwidth]{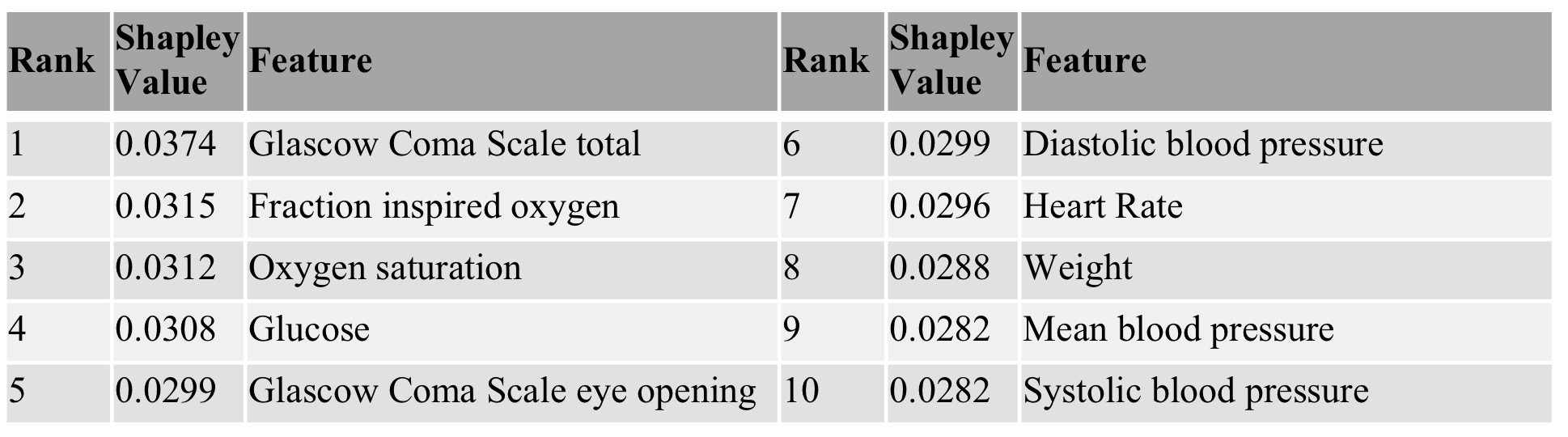}
\caption{ Top 10 Features of Clinical Variables based on Shapley Value.}
\label{fig:ehr_shapleyvalue}
\end{figure}

Next, we implement Shapley values to rank the important clinical variables. Shapley values\citep{lundberg2017unified} involve a game theory-based approach to explain the prediction of deep learning models. They measure the contribution of a given feature value to the difference from the actual prediction to the mean prediction. The top 10 out of 17 clinical variables (Figure \ref{fig:ehr_shapleyvalue}) show that for structured EHR data, the highest ranked variables also correlate with disease severity and poorer prognosis. These variables represent clinically important information such as mental status using the Glasgow Coma Scale, respiratory status and oxygenation, and hemodynamic measurements. They also provide interpretability of the directionality of impact for continuous variables, with poor prognostic variables like higher need for supplemental oxygen (Fraction inspired oxygen) increasing the likelihood for predicting  death, and favorable prognostic variables, like higher blood systolic and blood pressure decreasing the likelihood of predicting death.

\subsection{Discussion}
Vast clinical datasets provide the opportunity for deep learning techniques to study the problem of in-hospital mortality prediction. Compared to previous related work, which mostly considers single modality or only naively concatenates embeddings from different modalities, our work demonstrates a novel way to integrate multimodal knowledge and leverage clinical notes information for better predictions. To our best knowledge, this is the first work utilizing a transformer block to fuse clinical notes and clinical variable information while dealing with time series data in EHR data. We also conduct comprehensive experiments to demonstrate that our proposed method outperforms other methods by achieving high performance (AUCPR: 0.538, AUCROC: 0.877, F1:0.490).

The way we project two different modalities, unstructured clinical notes and structured clinical variables, into a universal shareable space with a transformer block is very efficient to leverage clinical notes and integrate clinical information. Meanwhile, the novel application of transformers on clinical data enables the model to consider information from all other time stamps when fusing the multimodal information because of the unique attention mechanism in the transformer block.

The ablation study of domain adaptive pretraining and task adaptive fine-tuning on various BERTs verifies the significance of pretraining and fine-tuning when we implement the BERT models on natural language text, especially on domain-specific clinical notes.

The analysis and visualization of important words in clinical notes also provide interesting findings. The ranking of words by IG values provides face validity that many of the important words used for prediction are clinically related to diseases or processes that are prognostically important, such as severity of respiratory disease or mental status. Other words, such as “care,” may be used in multiple contexts, and are more difficult to interpret as isolated words. Lastly, the clinical meaning of multiple words can change significantly with negation, such as “crackles” indicating abnormal lung exam findings, and “not crackles” indicating a normal lung exam. In the future, we will employ techniques like the NegEx algorithm\citep{chapman2001simple} to consider negation of  key words to better explain the clinical words' meaning.

\subsection{Conclusion}
In this paper, we demonstrate a novel transformer based model, Multimodal Transformer, to leverage clinical notes and fuse multimodal knowledge from clinical data. To our knowledge, we are the first to implement a transformer block to integrate both clinical notes and clinical variables while considering the time series information. The results demonstrate that our proposed Multimodal Transformer outperforms other methods. Additionally, we conduct different studies to further investigate the importance of domain adaptive pretraining and task adaptive fine-tuning for the Clinical BERTs. We also provide methods to interpret and visualize the important words in clinical notes using IG and Shapley methods, which demonstrate interesting findings on important features in clinical variables.

\bibliography{iclr2023_conference}
\bibliographystyle{iclr2023_conference}

% \appendix
% \section{Appendix}
% You may include other additional sections here.

\end{document}

%% file: math_commands.tex
\usepackage{amsmath,amsfonts,bm}

\def\eqref#1{equation~\ref{#1}}
\def\1{\bm{1}}

\DeclareMathAlphabet{\mathsfit}{\encodingdefault}{\sfdefault}{m}{sl}
\SetMathAlphabet{\mathsfit}{bold}{\encodingdefault}{\sfdefault}{bx}{n}

\newcommand{\R}{\mathbb{R}}

%% file: iclr2023_conference.bbl
\begin{thebibliography}{34}
\providecommand{\natexlab}[1]{#1}
\providecommand{\url}[1]{\texttt{#1}}
\expandafter\ifx\csname urlstyle\endcsname\relax
  \providecommand{\doi}[1]{doi: #1}\else
  \providecommand{\doi}{doi: \begingroup \urlstyle{rm}\Url}\fi

\bibitem[Alsentzer et~al.(2019)Alsentzer, Murphy, Boag, Weng, Jin, Naumann, and
  McDermott]{alsentzer2019publicly}
Emily Alsentzer, John~R Murphy, Willie Boag, Wei-Hung Weng, Di~Jin, Tristan
  Naumann, and Matthew McDermott.
\newblock Publicly available clinical bert embeddings.
\newblock \emph{arXiv preprint arXiv:1904.03323}, 2019.

\bibitem[Cai et~al.(2016)Cai, Perez-Concha, Coiera, Martin-Sanchez, Day, Roffe,
  and Gallego]{cai2016real}
Xiongcai Cai, Oscar Perez-Concha, Enrico Coiera, Fernando Martin-Sanchez,
  Richard Day, David Roffe, and Blanca Gallego.
\newblock Real-time prediction of mortality, readmission, and length of stay
  using electronic health record data.
\newblock \emph{Journal of the American Medical Informatics Association},
  23\penalty0 (3):\penalty0 553--561, 2016.

\bibitem[Chapman et~al.(2001)Chapman, Bridewell, Hanbury, Cooper, and
  Buchanan]{chapman2001simple}
Wendy~W Chapman, Will Bridewell, Paul Hanbury, Gregory~F Cooper, and Bruce~G
  Buchanan.
\newblock A simple algorithm for identifying negated findings and diseases in
  discharge summaries.
\newblock \emph{Journal of biomedical informatics}, 34\penalty0 (5):\penalty0
  301--310, 2001.

\bibitem[Dalal et~al.(2021)Dalal, Piniella, Fuller, Pong, Pardo, Bessa, Yoon,
  Lipsitz, and Schnipper]{dalal2021evaluation}
Anuj~K Dalal, Nicholas Piniella, Theresa~E Fuller, Denise Pong, Michael Pardo,
  Nathaniel Bessa, Catherine Yoon, Stuart Lipsitz, and Jeffrey~L Schnipper.
\newblock Evaluation of electronic health record-integrated digital health
  tools to engage hospitalized patients in discharge preparation.
\newblock \emph{Journal of the American Medical Informatics Association},
  28\penalty0 (4):\penalty0 704--712, 2021.

\bibitem[Devlin et~al.(2019)Devlin, Chang, Lee, and Toutanova]{devlin2019bert}
Jacob Devlin, Ming-Wei Chang, Kenton Lee, and Kristina Toutanova.
\newblock Bert: Pre-training of deep bidirectional transformers for language
  understanding.
\newblock In \emph{NAACL-HLT (1)}, 2019.

\bibitem[Deznabi et~al.(2021)Deznabi, Iyyer, and
  Fiterau]{deznabi2021predicting}
Iman Deznabi, Mohit Iyyer, and Madalina Fiterau.
\newblock Predicting in-hospital mortality by combining clinical notes with
  time-series data.
\newblock In \emph{Findings of the Association for Computational Linguistics:
  ACL-IJCNLP 2021}, pp.\  4026--4031, 2021.

\bibitem[Dong et~al.(2019)Dong, Rashidian, Wang, Hajagos, Zhao, Rosenthal,
  Kong, Saltz, Saltz, and Wang]{dong2019machine}
Xinyu Dong, Sina Rashidian, Yu~Wang, Janos Hajagos, Xia Zhao, Richard~N
  Rosenthal, Jun Kong, Mary Saltz, Joel Saltz, and Fusheng Wang.
\newblock Machine learning based opioid overdose prediction using electronic
  health records.
\newblock In \emph{AMIA Annual Symposium Proceedings}, volume 2019, pp.\  389.
  American Medical Informatics Association, 2019.

\bibitem[Gu et~al.(2021)Gu, Tinn, Cheng, Lucas, Usuyama, Liu, Naumann, Gao, and
  Poon]{gu2021domain}
Yu~Gu, Robert Tinn, Hao Cheng, Michael Lucas, Naoto Usuyama, Xiaodong Liu,
  Tristan Naumann, Jianfeng Gao, and Hoifung Poon.
\newblock Domain-specific language model pretraining for biomedical natural
  language processing.
\newblock \emph{ACM Transactions on Computing for Healthcare (HEALTH)},
  3\penalty0 (1):\penalty0 1--23, 2021.

\bibitem[Gururangan et~al.(2020)Gururangan, Marasovi{\'c}, Swayamdipta, Lo,
  Beltagy, Downey, and Smith]{gururangan2020don}
Suchin Gururangan, Ana Marasovi{\'c}, Swabha Swayamdipta, Kyle Lo, Iz~Beltagy,
  Doug Downey, and Noah~A Smith.
\newblock Don't stop pretraining: adapt language models to domains and tasks.
\newblock \emph{arXiv preprint arXiv:2004.10964}, 2020.

\bibitem[Harutyunyan et~al.(2019)Harutyunyan, Khachatrian, Kale, Ver~Steeg, and
  Galstyan]{harutyunyan2019multitask}
Hrayr Harutyunyan, Hrant Khachatrian, David~C Kale, Greg Ver~Steeg, and Aram
  Galstyan.
\newblock Multitask learning and benchmarking with clinical time series data.
\newblock \emph{Scientific data}, 6\penalty0 (1):\penalty0 1--18, 2019.

\bibitem[Henry et~al.(2016)Henry, Pylypchuk, Searcy, and
  Patel]{henry2016adoption}
J~Henry, Yuriy Pylypchuk, Talisha Searcy, and Vaishali Patel.
\newblock Adoption of electronic health record systems among us non-federal
  acute care hospitals: 2008--2015.
\newblock \emph{ONC data brief}, 35\penalty0 (35):\penalty0 2008--2015, 2016.

\bibitem[Huang et~al.(2019)Huang, Altosaar, and
  Ranganath]{huang2019clinicalbert}
Kexin Huang, Jaan Altosaar, and Rajesh Ranganath.
\newblock Clinicalbert: Modeling clinical notes and predicting hospital
  readmission.
\newblock \emph{arXiv preprint arXiv:1904.05342}, 2019.

\bibitem[Huang et~al.(2020)Huang, Pareek, Seyyedi, Banerjee, and
  Lungren]{huang2020fusion}
Shih-Cheng Huang, Anuj Pareek, Saeed Seyyedi, Imon Banerjee, and Matthew~P
  Lungren.
\newblock Fusion of medical imaging and electronic health records using deep
  learning: a systematic review and implementation guidelines.
\newblock \emph{NPJ digital medicine}, 3\penalty0 (1):\penalty0 1--9, 2020.

\bibitem[Johnson et~al.(2016)Johnson, Pollard, Shen, Lehman, Feng, Ghassemi,
  Moody, Szolovits, Anthony~Celi, and Mark]{johnson2016mimic}
Alistair~EW Johnson, Tom~J Pollard, Lu~Shen, Li-wei~H Lehman, Mengling Feng,
  Mohammad Ghassemi, Benjamin Moody, Peter Szolovits, Leo Anthony~Celi, and
  Roger~G Mark.
\newblock Mimic-iii, a freely accessible critical care database.
\newblock \emph{Scientific data}, 3\penalty0 (1):\penalty0 1--9, 2016.

\bibitem[Khadanga et~al.(2019)Khadanga, Aggarwal, Joty, and
  Srivastava]{khadanga2019using}
Swaraj Khadanga, Karan Aggarwal, Shafiq Joty, and Jaideep Srivastava.
\newblock Using clinical notes with time series data for icu management.
\newblock \emph{arXiv preprint arXiv:1909.09702}, 2019.

\bibitem[Kong et~al.(2020)Kong, Lin, and Hu]{kong2020using}
Guilan Kong, Ke~Lin, and Yonghua Hu.
\newblock Using machine learning methods to predict in-hospital mortality of
  sepsis patients in the icu.
\newblock \emph{BMC medical informatics and decision making}, 20\penalty0
  (1):\penalty0 1--10, 2020.

\bibitem[Lee et~al.(2020)Lee, Yoon, Kim, Kim, Kim, So, and
  Kang]{lee2020biobert}
Jinhyuk Lee, Wonjin Yoon, Sungdong Kim, Donghyeon Kim, Sunkyu Kim, Chan~Ho So,
  and Jaewoo Kang.
\newblock Biobert: a pre-trained biomedical language representation model for
  biomedical text mining.
\newblock \emph{Bioinformatics}, 36\penalty0 (4):\penalty0 1234--1240, 2020.

\bibitem[Li et~al.(2021)Li, Xin, Zhang, Fu, Zhou, and Lian]{li2021prediction}
Fuhai Li, Hui Xin, Jidong Zhang, Mingqiang Fu, Jingmin Zhou, and Zhexun Lian.
\newblock Prediction model of in-hospital mortality in intensive care unit
  patients with heart failure: machine learning-based, retrospective analysis
  of the mimic-iii database.
\newblock \emph{BMJ open}, 11\penalty0 (7):\penalty0 e044779, 2021.

\bibitem[Li et~al.(2020)Li, Rao, Solares, Hassaine, Ramakrishnan, Canoy, Zhu,
  Rahimi, and Salimi-Khorshidi]{li2020behrt}
Yikuan Li, Shishir Rao, Jos{\'e} Roberto~Ayala Solares, Abdelaali Hassaine,
  Rema Ramakrishnan, Dexter Canoy, Yajie Zhu, Kazem Rahimi, and Gholamreza
  Salimi-Khorshidi.
\newblock Behrt: transformer for electronic health records.
\newblock \emph{Scientific reports}, 10\penalty0 (1):\penalty0 1--12, 2020.

\bibitem[Lundberg \& Lee(2017)Lundberg and Lee]{lundberg2017unified}
Scott~M Lundberg and Su-In Lee.
\newblock A unified approach to interpreting model predictions.
\newblock \emph{Advances in neural information processing systems}, 30, 2017.

\bibitem[Paszke et~al.(2019)Paszke, Gross, Massa, Lerer, Bradbury, Chanan,
  Killeen, Lin, Gimelshein, Antiga, et~al.]{paszke2019pytorch}
Adam Paszke, Sam Gross, Francisco Massa, Adam Lerer, James Bradbury, Gregory
  Chanan, Trevor Killeen, Zeming Lin, Natalia Gimelshein, Luca Antiga, et~al.
\newblock Pytorch: An imperative style, high-performance deep learning library.
\newblock \emph{Advances in neural information processing systems}, 32, 2019.

\bibitem[Rahman et~al.(2020)Rahman, Hasan, Lee, Zadeh, Mao, Morency, and
  Hoque]{rahman2020integrating}
Wasifur Rahman, Md~Kamrul Hasan, Sangwu Lee, Amir Zadeh, Chengfeng Mao,
  Louis-Philippe Morency, and Ehsan Hoque.
\newblock Integrating multimodal information in large pretrained transformers.
\newblock In \emph{Proceedings of the conference. Association for Computational
  Linguistics. Meeting}, volume 2020, pp.\  2359. NIH Public Access, 2020.

\bibitem[Ramachandram \& Taylor(2017)Ramachandram and
  Taylor]{ramachandram2017deep}
Dhanesh Ramachandram and Graham~W Taylor.
\newblock Deep multimodal learning: A survey on recent advances and trends.
\newblock \emph{IEEE signal processing magazine}, 34\penalty0 (6):\penalty0
  96--108, 2017.

\bibitem[Rocheteau et~al.(2021)Rocheteau, Tong, Veli{\v{c}}kovi{\'c}, Lane, and
  Li{\`o}]{rocheteau2021predicting}
Emma Rocheteau, Catherine Tong, Petar Veli{\v{c}}kovi{\'c}, Nicholas Lane, and
  Pietro Li{\`o}.
\newblock Predicting patient outcomes with graph representation learning.
\newblock \emph{arXiv preprint arXiv:2101.03940}, 2021.

\bibitem[Schwartz et~al.(2021)Schwartz, Moy, Rossetti, Elhadad, and
  Cato]{schwartz2021clinician}
Jessica~M Schwartz, Amanda~J Moy, Sarah~C Rossetti, No{\'e}mie Elhadad, and
  Kenrick~D Cato.
\newblock Clinician involvement in research on machine learning--based
  predictive clinical decision support for the hospital setting: A scoping
  review.
\newblock \emph{Journal of the American Medical Informatics Association},
  28\penalty0 (3):\penalty0 653--663, 2021.

\bibitem[Sheikhalishahi et~al.(2020)Sheikhalishahi, Balaraman, and
  Osmani]{sheikhalishahi2020benchmarking}
Seyedmostafa Sheikhalishahi, Vevake Balaraman, and Venet Osmani.
\newblock Benchmarking machine learning models on multi-centre eicu critical
  care dataset.
\newblock \emph{Plos one}, 15\penalty0 (7):\penalty0 e0235424, 2020.

\bibitem[Si et~al.(2021)Si, Du, Li, Jiang, Miller, Wang, Zheng, and
  Roberts]{si2021deep}
Yuqi Si, Jingcheng Du, Zhao Li, Xiaoqian Jiang, Timothy Miller, Fei Wang, W~Jim
  Zheng, and Kirk Roberts.
\newblock Deep representation learning of patient data from electronic health
  records (ehr): A systematic review.
\newblock \emph{Journal of Biomedical Informatics}, 115:\penalty0 103671, 2021.

\bibitem[Sundararajan et~al.(2017)Sundararajan, Taly, and
  Yan]{sundararajan2017axiomatic}
Mukund Sundararajan, Ankur Taly, and Qiqi Yan.
\newblock Axiomatic attribution for deep networks.
\newblock In \emph{International conference on machine learning}, pp.\
  3319--3328. PMLR, 2017.

\bibitem[Teixeira et~al.(2017)Teixeira, Wei, Cronin, Mo, VanHouten, Carroll,
  LaRose, Bastarache, Rosenbloom, Edwards, et~al.]{teixeira2017evaluating}
Pedro~L Teixeira, Wei-Qi Wei, Robert~M Cronin, Huan Mo, Jacob~P VanHouten,
  Robert~J Carroll, Eric LaRose, Lisa~A Bastarache, S~Trent Rosenbloom, Todd~L
  Edwards, et~al.
\newblock Evaluating electronic health record data sources and algorithmic
  approaches to identify hypertensive individuals.
\newblock \emph{Journal of the American Medical Informatics Association},
  24\penalty0 (1):\penalty0 162--171, 2017.

\bibitem[Teo et~al.(2021)Teo, Yong, Chuah, Hum, Tee, Xia, and
  Lai]{teo2021current}
Kareen Teo, Ching~Wai Yong, Joon~Huang Chuah, Yan~Chai Hum, Yee~Kai Tee,
  Kaijian Xia, and Khin~Wee Lai.
\newblock Current trends in readmission prediction: An overview of approaches.
\newblock \emph{Arabian journal for science and engineering}, pp.\  1--18,
  2021.

\bibitem[Wolf et~al.(2019)Wolf, Debut, Sanh, Chaumond, Delangue, Moi, Cistac,
  Rault, Louf, Funtowicz, et~al.]{wolf2019huggingface}
Thomas Wolf, Lysandre Debut, Victor Sanh, Julien Chaumond, Clement Delangue,
  Anthony Moi, Pierric Cistac, Tim Rault, R{\'e}mi Louf, Morgan Funtowicz,
  et~al.
\newblock Huggingface's transformers: State-of-the-art natural language
  processing.
\newblock \emph{arXiv preprint arXiv:1910.03771}, 2019.

\bibitem[Yang \& Wu(2021)Yang and Wu]{yang2021leverage}
Bo~Yang and Lijun Wu.
\newblock How to leverage multimodal ehr data for better medical predictions?
\newblock \emph{arXiv preprint arXiv:2110.15763}, 2021.

\bibitem[Zhang et~al.(2020)Zhang, Yin, Zeng, Yuan, and
  Zhang]{zhang2020combining}
Dongdong Zhang, Changchang Yin, Jucheng Zeng, Xiaohui Yuan, and Ping Zhang.
\newblock Combining structured and unstructured data for predictive models: a
  deep learning approach.
\newblock \emph{BMC medical informatics and decision making}, 20\penalty0
  (1):\penalty0 1--11, 2020.

\bibitem[Zheng et~al.(2017)Zheng, Lu, Ghasemzadeh, Hayek, Quyyumi, Wang,
  et~al.]{zheng2017effective}
Shuai Zheng, James~J Lu, Nima Ghasemzadeh, Salim~S Hayek, Arshed~A Quyyumi,
  Fusheng Wang, et~al.
\newblock Effective information extraction framework for heterogeneous clinical
  reports using online machine learning and controlled vocabularies.
\newblock \emph{JMIR medical informatics}, 5\penalty0 (2):\penalty0 e7235,
  2017.

\end{thebibliography}
